# THE EXPERT SYSTEM DESIGNED TO IMPROVE CUSTOMER SATISFACTION

P.Isakki alias Devi[1] and Dr.S.P.Rajagopalan[2]

[1]Research Scholar, Vels University, Chennai, India
akilmira@yahoo.com
[2]GKM College of Engineering & Technology, Chennai, India
sasirekaraj@yahoo.co.in

## ABSTRACT

*Customer Relationship Management becomes a leading business strategy in highly competitive business environment. It aims to enhance the performance of the businesses by improving the customer satisfaction and loyalty. The objective of this paper is to improve customer satisfaction on product's colors and design with the help of the expert system developed by using Artificial Neural Networks. The expert system's role is to capture the knowledge of the experts and the data from the customer requirements, and then, process the collected data and form the appropriate rules for choosing product's colors and design. In order to identify the hidden pattern of the customer's needs, the Artificial Neural Networks technique has been applied to classify the colors and design based upon a list of selected information. Moreover, the expert system has the capability to make decisions in ranking the scores of the colors and design presented in the selection. In addition, the expert system has been validated with a different customer types.*

## 1. INTRODUCTION

The core part of CRM activities is to understand customer requirements and retain profitable customers. To reach it in a highly competitive market, satisfying customer's needs is the key to business success [1]. Unprecedented growth of competition has raised the importance of retaining current customers. Retaining existing customers is much less expensive and difficult than recruiting new customers in a mature market. So customer retention is a significant stage in Customer Relation Management, which is also the most important growth point of profit [13].

Factors that influence customer satisfaction degree are concerned by all enterprise managers [9]. Marketing literature states that it is more costly to engage a new customer than to retain an existing loyal customer. Churn prediction models are developed by academics and practitioners to effectively manage and control customer churn in order to retain existing customers [14]. So, Customer satisfaction is most important.

Data mining (DM) methodology has a tremendous contribution for researchers to extract the hidden knowledge and information which have been inherited in the data used by researchers [7]. Data mining has a tremendous contribution to the extraction of knowledge and information which have been hidden in a large volume of data [8]. The concept of customer satisfaction and loyalty (CS&L) has attracted much attention in recent years. A key motivation for the fast growing emphasis on CS&L can be attributed to the fact that higher customer satisfaction and loyalty can lead to stronger competitive position resulting in larger market share and profitability [6].

However, it is a difficult and a complex task to identify the customer's needs such as colors and design of the products. The objective of this paper is to design and implement the expert system in order to assess customer satisfaction and reveal appropriate strategies to improve it. As the customer satisfaction on colors and design can have a complex hidden pattern and, therefore, the approach of the paper should have an ability to perform pattern recognition, classification and forecast which make the artificial neural networks an appropriate technique to be applied in the expert system. The conceptual work of the paper is illustrated in Figure 1, in which the assumption of the customer requirements and expert system are based upon the statement that ''in general, the same customer group will like the same colors''.

A vast variety of colors mixing in different products that makes it a difficult and complicated task to identify the customer's needs. The contribution of this paper is in designing the system that is the combination of the expert system and the ANN. The customers can interact with the interface of the expert system to ask and get the advices from the system. Correlation Coefficient can be found. According to that, we can identify the customer's behaviour.

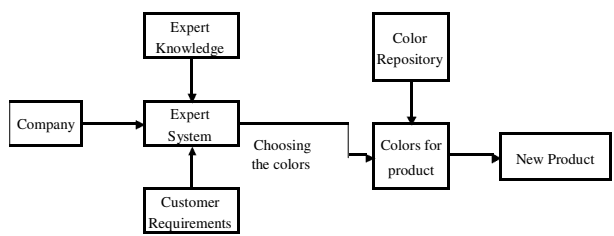

Figure 1. The Conceptual Work

## 2. RELATED WORKS

The expert system's role is in the preparation to capture the knowledge of the experts and the data from the customer's requirements. The system has the capability to compile the collected data and form the appropriate rules for choosing fragrance notes for the products. In order to identify the hidden pattern of the customer's needs, the artificial neural networks technique has been applied to classify the fragrance notes based upon a list of selected information [3].

The expert system's role is in the preparation to capture the data from the customer's requirements and predict appropriate perfume. For this end, factors of perfume costumers' decision were recognized using Fuzzy Delphi method and a back propagation neural network classification model was developed and trained with 2303 data of customers [4].

The proposed business intelligent system for demand forecasting proves to give more accurate prediction for future demands compared to the existing models and practices in spare parts inventory management. This helps inventory managers to better manage their supply chain performance by reducing reaching days and service level simultaneously. Reaching day as a measure of inventory level is generally reduced successfully by the retailers at the cost of service level in most of the places [5].

## 3. EXPERT SYSTEM AND ARTIFICIAL NEURAL NETWORKS

### 3.1 Expert System

An expert system is the computer system that emulates the behaviour of human experts in a well-specified manner, and narrowly defines the domain of knowledge. It captures the knowledge and heuristics that an expert employs in a specific task. An overview of current technologies applied with an expert system that is developed for Database Management System, Decision Support System, and the other intelligent systems such as Neural Networks System, Genetic Algorithm, etc., can be seen in [2].

An Expert System is viewed as a computer simulation of a human expert. Expert Systems are an emerging technology with many areas of potential applications. Expert systems are artificial intelligence (AI) tools that capture the expertise of knowledge workers and provide advice to (usually) non-experts in a given domain. Thus, expert systems constitute a subset of the class of AI systems primarily concerned with transferring knowledge from experts to novices. Knowledge-based expert systems, or simply expert systems, use human knowledge to solve problems that normally would require human intelligence. These expert systems represent the expertise knowledge as data or rules within the computer. These rules and data can be called upon when needed to solve problems. There are three main parts to the expert system: knowledge base, a set of if-then rules; working memory, a database of facts; inference engine, the reasoning logic to create rules and data.

In this paper, the expert system roles have been designed to capture the knowledge of the experts and the data of customer requirements. After that the system is able to process these data and form rules to customize colors in products. Finally, the expert system will deliver the decision ranking of the scores based upon the product samples, which are presented in the selection process.

### 3.2 Artificial Neural Networks

Artificial Neural Networks (ANNs) are generic non-linear function approximations extensively used for pattern recognition and classification. An artificial neural network (ANN), usually called neural network (NN), is a mathematical model or computational model that is inspired by the structure and/or functional aspects of biological neural networks. A neural network consists of an interconnected group of artificial neurons, and it processes information using a connectionist approach to computation. In most cases an ANN is an adaptive system that changes its structure based on external or internal information that flows through the network during the learning phase.

Modern neural networks are non-linear statistical data modelling tools. They are usually used to model complex relationships between inputs and outputs or to find patterns in data. An ANN is a collection of basic units, called neurons, computing a non-linear function of their input. Every input has an assigned weight that determines the impact this input has on the overall output of the node. In Figure 2 it is possible to see a schematic representation of such an artificial neuron, where $w_{ji}$ is the weight of the connection from neuron i to neuron j, and $o_j$ is the activation or output of neuron j. Unit j output is obtained by ideally following a two step-procedure. First the total weighted input $z_j$ is computed using the formula $z_j = \sum_i w_{ji} o_i$ where $o_i$ is the activity level of the i-th unit in the previous layer and $w_{ji}$ is the weight of the connection between the i-th and the j-th unit. Then, the neuron output is obtained as a non-linear function, shown in figure 2.

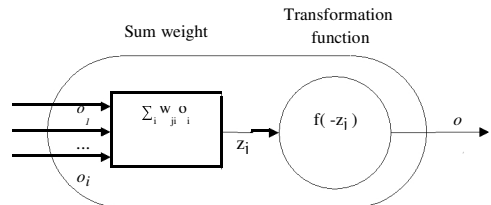

Figure 2. Artificial neuron *j*.

By interconnecting a proper number of nodes in a suitable way and by setting the weights to appropriate values, ANN can approximate any non-linear function with arbitrary precision. This structure of nodes and connections, known as network topology, together with the weights of the connections, determines the final behaviour of the network. Given ANN topology and a training set, it is possible to optimize the values of the weights in order to minimize an error function by means of any back propagation algorithm, standard optimization techniques or randomized algorithms.

Upward and downward trends in sales signify new market trends and understanding of sales trends is important for marketing as well as for customer retention [10]. To estimate the neural network weights of their neural network models, backward propagation (BP) is useful [11]. Neural networks can be very successful in learning and estimating data analysis, data pre-processing and training [12].

In practical terms, a product may have several colors mixed together. Based upon this complicated situation, we need to apply the artificial intelligence system by using ANN to classify the colors and design that are appropriate to the consumer requirements. In this paper, we use the multilayer architecture, back propagation algorithm, in a supervised learning method, to classify the colors that are preferred by consumers.

## 4. PROPOSED DESIGN

In order to develop the expert system, certain steps, which consist of gathering the raw data, system designing, programming and testing, verifying and validating, and evaluation must be fulfilled.

### 4.1. The Expert System Design

Based upon Medsker and Liebowitz concerning the theory of expert systems, which should have three main components that are:

- ➢ User interface
- ➢ Inference engine for making decision
- ➢ Database for storing the data rules, and training the system.

Figure 3 illustrates the system model that embraces the ANN in the expert system. The ANN is preferred in this paper because its ability to clarify and classify complex hidden pattern of the customer requirements and to forecast colors and design that can satisfy different types of customer.

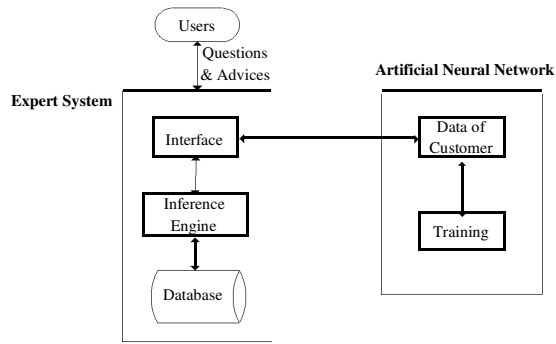

Figure 3. The Expert System Model

The customers can interact with the interface of the expert system to ask and get the advices from the system. The Inference engine consists of:
- The production rules from expert knowledge
- The data of customers that are classified by the ANN system.

## 4.2 The Artificial Neural Networks Design

The commonest type of artificial neural network consists of three groups, or layers, of units: a layer of "**input**" units is connected to a layer of "**hidden**" units, which is connected to a layer of **"output"** units.

An artificial neuron is a multi input and single output device having two operation modes- the training mode and the using mode. The neuron is trained to fire (or not), for particular input patterns, in the training mode. In the using mode, when a taught input pattern is detected at the input, its associated output becomes the current output. If the input pattern does not belong in the taught list of input patterns, the firing rule is used to determine whether to fire or not.

The activity of the input units represents the raw information that is fed into the network. The activity of each hidden unit is determined by the activities of the input units and the weights on the connections between the input and the hidden units. The behaviour of the output units depends on the activity of the hidden units and the weights between the hidden and output units.

Medsker and Liebowitz mention that the common multilayer architecture is more appropriate for classification applications, and this is the reason we select multilayer architecture to classify colors and design. They recommend including eight input nodes (which means eight types of customer) in the first layer. The second layer (called hidden layer) should have 30 nodes. The third layer (called the output layer) should have 52 nodes (which means 52 types of mixed colors). So, the total multilayer architecture has 8 input nodes, 30 hidden nodes, and 52 output nodes.

We choose the mathematical model at a neuron as shown in Figure. 2. Considering the task to train the data of customer requirements, we apply the multilayer architecture with back propagation learning rule by adjusting the weight matrix. The structure of the ANN is illustrated in Figure 4.

Phase 1: Propogation
Each propagation involves the following steps:

Forward propagation of a training pattern's input through the neural network in order to generate the propagation's output activations.
Back propagation of the propagation's output activations through the neural network using the training pattern's target in order to generate the deltas of all output and hidden neurons.

Phase 2: Weight Update:

For each weight-synapse:
> Multiply its output delta and input activation to get the gradient of the weight.
> Bring the weight in the opposite direction of the gradient by subtracting a ratio of it from the weight.

This ratio influences the speed and quality of learning; it is called the learning rate. The sign of the gradient of a weight indicates where the error is increasing, this is why the weight must be updated in the opposite direction.

Repeat the phase 1 and 2 until the performance of the network is good enough.

The model has the mathematic functions as below, and the model parameters are illustrated in Table 1.

Transformation function:

$$o_j = f(z_j) = 1/1 + \exp(-z_j) \qquad z_j = \sum_i w_{ji} o_i \qquad (1)$$

- The error equation:

$$e_j(n) = d_j(n) - o_j(n) \qquad (2)$$

- Slope at output layer:

$$\delta_j^{(L)}(n) = e_j(n) o_j(n)[1 - o_j(n)] \qquad (3)$$

- Slope at the other layer:

$$\delta_j^{(l)}(n) = y_j(n)[1 - y_j(n)] \sum_k \delta_k^{(l+1)}(n) w_{kj}^{(l+1)}(n) \qquad (4)$$

- Weight update:

$$w_{ji}^{(l)}(n+1) = w_{ji}^{(l)}(n) + \alpha[w_{ji}^{(l)}(n) - w_{ji}^{(l)}(n-1)] + \eta \delta_j^{(l)}(n) y_j^{(l-1)}(n) \qquad (5)$$

- Learning rate ($\alpha$)=0.2 (according to [3]).

- Momentum constant ($\eta$)=0.5 (according to [3]).

From Equation (1) we can illustrate the ANN computation as follows.
*Input layer*: The input signal in each node is ''1'' and the node that has no signal is ''0''.

*Hidden layer*: At a node, the input values by the weights are sum to this node and transfer to next layer by the Equation (1).

Table 1. Model parameters

| | |
|---|---|
| $y$: Output at a layer | $L$: Output layer |
| $d$: Desired output | $l$: First and second (hidden) layer |
| $e$: Error | $n$: Learning iteration (epoch) |
| $o$: Computation output | $\delta_j$: Slope at a layer |
| $\alpha$: Learning rate | $w_{ji}$: Weight matrix |
| $\eta$: Momentum constant | $\delta_k$: Previous slope (back propagate) |

From Equation (2) to (5) when we try to train the system, the weight matrix will be changed, and the networks computation will classify the data more precisely than the previous computation. It has been computed in order to achieve the result and the error can be computed by Equation (2).

Output layer: Then, the error values are propagated to the previous layer to adjust the weight matrix. At first we calculate the slope at a layer by the Equation (3). The weights updated are computed by the Equation (5).

Hidden and input layer: The hidden and input layer use similar computing, but the Equation (4) is substitute to the Equation (3). Ultimately, the hidden layer can precisely transfer the data from the output layer to the first layer.

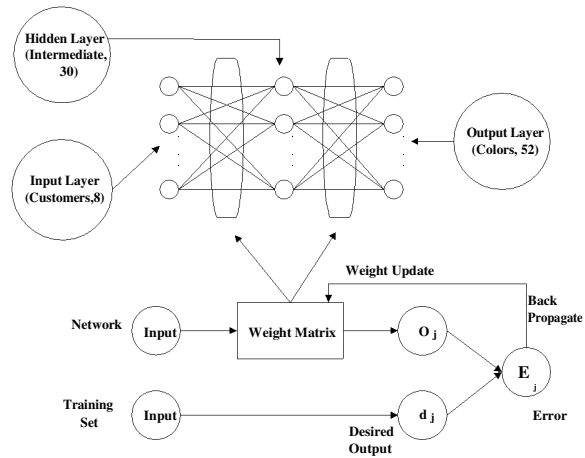

Figure 4. The multilayer architecture with back propagation.

## 5. EVALUATING DATA

Our aim is to produce a new product for young male / female as target groups. The firm needs to select some colors and design from their repository based upon the ability of the expert system.

Our aim is to watch the chosen colors and design advised by the system, for each type of customer.

According to [2], the evaluating data should consist of 16 cases as follows:

- 2 decision, by expert knowledge and data of customer requirements,
- 2 genders, male, and female,
- 4 age groups, teen, young, adult, and senior.

**5.1 Gathering the Raw Data**

Gathering the raw data contains expert knowledge acquisition and the customer requirements data. We have to acquire knowledge directly from the products. The acquired knowledge consists of the fundamental principles of using colors, the materials, the techniques, how colors are used in producing, and the criteria for choosing them that are appropriate to the consumers. This means the data in this viewpoint is from the expert and manufacturer side. The customer requirements data, as mentioned earlier, the first layer, the input layer, should have eight nodes, the list below shows eight varieties of customers.

- Male–teen (13–19 years old)
- Male–young (20–29 years old)
- Male–adult (30–45 years old)
- Male–senior (over 46 years old)
- Female–teen (13–19 years old)
- Female–young (20–29 years old)
- Female–adult (30–45 years old)
- Female–senior (over 46 years old)

**5.2 Performance Evaluation**

After completing the recording of the expert knowledge and the details of the eight groups of customers (male/female of teen, young, adult, and old) purchased into the expert system on chosen products' characteristics about colors and design. The aim of the paper is to evaluate the accuracy between the result advised by the expert system and the chosen product (colors / design / design) from each type of any customer. So, the higher number of customer satisfaction on the product (colors / design) advised by the expert system means higher accuracy of the expert system.

The expert system performance evaluation is illustrated in Table 2, for example, at product sample S6, 29 out of 47 females at a young age chose similar to the result advised from the expert system, which means it can predict precisely at approximate 61.7%.

Based upon the firm target groups of male/female at young age, the result advised from the expert system (higher than 60%) can satisfy the firm's management.

However, the average percentage of the correctness is 62.6% that is a moderate to high correct result. The rest of the correctness percentages are all above 50%.

Table 2.  Performance evaluation of Sample Data

| Product Samples | Male | | | | Female | | | | Total |
|---|---|---|---|---|---|---|---|---|---|
| | Teen | Young | Adult | Old | Teen | Young | Adult | Old | |
| S1 | 35 | 1 | 1 | 1 | 3 | 1 | 1 | 1 | **44** |
| S2 | 6 | 25 | 1 | 1 | 3 | 14 | 1 | 1 | **52** |
| S3 | 1 | 1 | 6 | 0 | 1 | 1 | 2 | 0 | **12** |
| S4 | 1 | 0 | 1 | 9 | 1 | 0 | 1 | 3 | **16** |
| S5 | 3 | 1 | 1 | 0 | 55 | 1 | 4 | 0 | **65** |
| S6 | 2 | 7 | 1 | 2 | 13 | 29 | 4 | 2 | **60** |
| S7 | 1 | 0 | 0 | 0 | 1 | 1 | 40 | 0 | **43** |
| S8 | 1 | 1 | 1 | 3 | 1 | 0 | 2 | 7 | **16** |
| **Total** | **50** | **36** | **12** | **16** | **78** | **47** | **55** | **14** | **308** |

| Samples | S1 | S2 | S3 | S4 | S5 | S6 | S7 | S8 | Average |
|---|---|---|---|---|---|---|---|---|---|
| **%Correct** | **70.0** | **69.4** | **50.0** | **56.3** | **70.5** | **61.7** | **72.7** | **50.0** | **62.6** |

Customer preferences can be easily analysed from this table. Purchased products from each sample can be identified easily. The list of male and female customers who purchased products has been given in table 3. The product wise list for each customer has been given in table 4.

Table 3.  Percentage of Male & Female customers purchased products

| Product Samples | Male | | | | Female | | | |
|---|---|---|---|---|---|---|---|---|
| | Teen | Young | Adult | Old | Teen | Young | Adult | Old |
| **S1** | 70.0 | 2.8 | 8.3 | 6.3 | 3.8 | 2.1 | 1.8 | 7.1 |
| **S2** | 12.0 | 69.4 | 8.3 | 6.3 | 3.8 | 29.8 | 1.8 | 7.1 |
| **S3** | 2.0 | 2.8 | 50.0 | 0.0 | 1.3 | 2.1 | 3.6 | 0.0 |
| **S4** | 2.0 | 0.0 | 8.3 | 56.3 | 1.3 | 0.0 | 1.8 | 21.4 |
| **S5** | 6.0 | 2.8 | 8.3 | 0.0 | 70.5 | 2.1 | 7.3 | 0.0 |
| **S6** | 4.0 | 19.4 | 8.3 | 12.5 | 16.7 | 61.7 | 7.3 | 14.3 |
| **S7** | 2.0 | 0.0 | 0.0 | 0.0 | 1.3 | 2.1 | 72.7 | 0.0 |
| **S8** | 2.0 | 2.8 | 8.3 | 18.8 | 1.3 | 0.0 | 3.6 | 50.0 |
| | | | | | | | | |
| **%** | **100** | **100** | **100** | **100** | **100** | **100** | **100** | **100** |

Table 4.  Percentage of Productwise list for each customer

| Types of Customers | S1 | S2 | S3 | S4 | S5 | S6 | S7 | S8 |
|---|---|---|---|---|---|---|---|---|
| **MaleTeen** | 79.5 | 11.5 | 8.3 | 6.3 | 4.6 | 3.3 | 2.3 | 6.3 |
| **MaleYoung** | 2.3 | 48.1 | 8.3 | 0.0 | 1.5 | 11.7 | 0.0 | 6.3 |
| **MaleAdult** | 2.3 | 1.9 | 50.0 | 6.3 | 1.5 | 11.7 | 0.0 | 6.3 |
| **MaleOld** | 2.3 | 1.9 | 0.0 | 56.3 | 0.0 | 3.3 | 0.0 | 18.0 |
| **FemaleTeen** | 6.8 | 5.8 | 8.3 | 6.3 | 84.6 | 21.7 | 2.3 | 6.3 |
| **FemaleYoung** | 2.3 | 26.9 | 8.3 | 0.0 | 1.5 | 48.3 | 2.3 | 0.0 |
| **FemaleAdult** | 2.3 | 1.9 | 16.7 | 6.3 | 6.2 | 6.7 | 93.0 | 12.5 |

| **FemaleOld** | 2.3 | 1.9 | 0.0 | 18.8 | 0.0 | 3.3 | 0.0 | 43.8 |
|---|---|---|---|---|---|---|---|---|
| **%** | **100** | **100** | **100** | **100** | **100** | **110** | **100** | **100** |

**5.3 Correlation Coefficient**

Correlation coefficient (r) is a measure of the degree of correlation between two quantities or variables, such as the rates of return on stocks and on bonds. The quantity r, called the linear correlation coefficient, measures the strength and the direction of a linear relationship between two variables. The mathematical formula for computing r is:

$$r = \frac{n\sum xy - (\sum x)(\sum y)}{\sqrt{n(\sum x^2) - (\sum x)^2} \sqrt{n(\sum y^2) - (\sum y)^2}}$$

where n is the number of pairs of data.

The value of r is such that -1 < r < +1. The + and – signs are used for positive linear correlations and negative linear correlations, respectively. If x and y have a strong positive linear correlation, r is close to +1. An r value of exactly +1 indicates a perfect positive. If x and y have a strong negative linear correlation, r is close to -1. An r value of exactly -1 indicates a perfect negative fit. If there is no linear correlation or a weak linear correlation, r is close to 0. A negative coefficient of correlation indicates an inverse or negative relationship, whereas a positive value indicates a direct or positive relationship.

The range of values is from -1 to +1 inclusive. A zero (0) value indicates that no correlation exists. Correlation coefficients are useful in asset class identification and portfolio diversification. The following figure 5 gives the details about correlation coefficient.

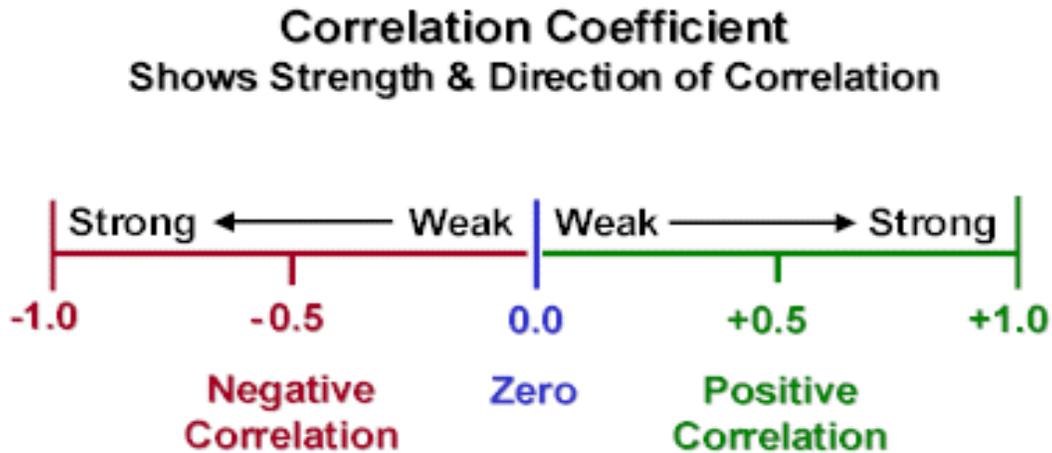

Figure 5. Correlation Coefficient

Advanced Computing: An International Journal ( ACIJ ), Vol.2, No.6, November 2011Table 5. Correlation Coefficients between Male & Female

|  | Teen | Young | Adult | Old |
|---|---|---|---|---|
| **Correlation Coefficient** | **-0.11** | **0.55** | **-0.33** | **0.52** |

**Correlations**

|  |  | M_TEEN | F_TEEN |
|---|---|---|---|
| M_TEEN | Pearson Correlation | 1 | -.110 |
|  | Sig. (2-tailed) | . | .795 |
|  | N | 8 | 8 |
| F_TEEN | Pearson Correlation | -.110 | 1 |
|  | Sig. (2-tailed) | .795 | . |
|  | N | 8 | 8 |

**Correlations**

|  |  | M_TEEN | M_YOUNG | M_ADULT | M_OLD | F_TEEN | F_YOUNG | F_ADULT | F_OLD |
|---|---|---|---|---|---|---|---|---|---|
| M_TEEN | Pearson Correlation | 1 | -.030 | -.138 | -.173 | -.110 | -.130 | -.215 | -.166 |
|  | Sig. (2-tailed) | . | .944 | .745 | .682 | .795 | .759 | .609 | .694 |
|  | N | 8 | 8 | 8 | 8 | 8 | 8 | 8 | 8 |
| M_YOUNG | Pearson Correlation | -.030 | 1 | -.117 | -.160 | -.111 | .548 | -.230 | -.112 |
|  | Sig. (2-tailed) | .944 | . | .783 | .706 | .793 | .159 | .584 | .792 |
|  | N | 8 | 8 | 8 | 8 | 8 | 8 | 8 | 8 |
| M_ADULT | Pearson Correlation | -.138 | -.117 | 1 | -.204 | -.144 | -.144 | -.330 | -.227 |
|  | Sig. (2-tailed) | .745 | .783 | . | .628 | .733 | .734 | .425 | .588 |
|  | N | 8 | 8 | 8 | 8 | 8 | 8 | 8 | 8 |
| M_OLD | Pearson Correlation | -.173 | -.160 | -.204 | 1 | -.283 | -.095 | -.299 | .517 |
|  | Sig. (2-tailed) | .682 | .706 | .628 | . | .498 | .823 | .472 | .189 |
|  | N | 8 | 8 | 8 | 8 | 8 | 8 | 8 | 8 |
| F_TEEN | Pearson Correlation | -.110 | -.111 | -.144 | -.283 | 1 | .015 | -.121 | -.303 |
|  | Sig. (2-tailed) | .795 | .793 | .733 | .498 | . | .972 | .775 | .465 |
|  | N | 8 | 8 | 8 | 8 | 8 | 8 | 8 | 8 |
| F_YOUNG | Pearson Correlation | -.130 | .548 | -.144 | -.095 | .015 | 1 | -.149 | -.053 |
|  | Sig. (2-tailed) | .759 | .159 | .734 | .823 | .972 | . | .726 | .900 |
|  | N | 8 | 8 | 8 | 8 | 8 | 8 | 8 | 8 |
| F_ADULT | Pearson Correlation | -.215 | -.230 | -.330 | -.299 | -.121 | -.149 | 1 | -.310 |
|  | Sig. (2-tailed) | .609 | .584 | .425 | .472 | .775 | .726 | . | .455 |
|  | N | 8 | 8 | 8 | 8 | 8 | 8 | 8 | 8 |
| F_OLD | Pearson Correlation | -.166 | -.112 | -.227 | .517 | -.303 | -.053 | -.310 | 1 |
|  | Sig. (2-tailed) | .694 | .792 | .588 | .189 | .465 | .900 | .455 | . |
|  | N | 8 | 8 | 8 | 8 | 8 | 8 | 8 | 8 |

81

**Correlations**

|  |  | M_YOUNG | F_YOUNG |
|---|---|---|---|
| M_YOUNG | Pearson Correlation | 1 | .548 |
|  | Sig. (2-tailed) | . | .159 |
|  | N | 8 | 8 |
| F_YOUNG | Pearson Correlation | .548 | 1 |
|  | Sig. (2-tailed) | .159 | . |
|  | N | 8 | 8 |

**Correlations**

|  |  | M_ADULT | F_ADULT |
|---|---|---|---|
| M_ADULT | Pearson Correlation | 1 | -.330 |
|  | Sig. (2-tailed) | . | .425 |
|  | N | 8 | 8 |
| F_ADULT | Pearson Correlation | -.330 | 1 |
|  | Sig. (2-tailed) | .425 | . |
|  | N | 8 | 8 |

**Correlations**

|  |  | M_OLD | F_OLD |
|---|---|---|---|
| M_OLD | Pearson Correlation | 1 | .517 |
|  | Sig. (2-tailed) | . | .189 |
|  | N | 8 | 8 |
| F_OLD | Pearson Correlation | .517 | 1 |
|  | Sig. (2-tailed) | .189 | . |
|  | N | 8 | 8 |

Table 5 gives the details about correlation coefficients between male and female customers. Correlation coefficient between Male teen and Female Teen is -0.11. It has negative correlation. Correlation coefficient between Male Young and Female Young is 0.55. It has positive correlation. Correlation coefficient between Male adult and Female adult is -0.33. It has negative correlation. Correlation coefficient between Male old and Female old is 0.52. It has positive correlation.

Table 6 lists the details about correlation coefficient between male and female for each products. Samples S1 & S2 contain correlation coefficient as 1. Sample S7 has -0.32. Sample S3, S4, S5, S6, S8 have the correlation coefficient as >=0.90.

Table 6. Correlation Coefficients between Male & Female for Each Products

| Samples | Correlation Coefficient |
|---|---|
| S1 | 1.00 |
| S2 | 1.00 |
| S3 | 0.90 |
| S4 | 0.96 |
| S5 | 0.94 |
| S6 | 0.93 |
| S7 | -0.32 |
| S8 | 0.96 |

## 6. CONCLUSIONS

Across market segments, a vast variety of colors mixing in different products that makes it a difficult and complicated task to identify the customer's needs. The contribution of this paper is in designing the system that is the combination of the expert system and the ANN. The customers can interact with the interface of the expert system to ask and get the advices from the system.

Another main result, knowledge from data analyzing, guides us in detail about how to utilize customer behaviour on preferred products (colors) in combination with knowledge captured from the expert in prediction of the colors. This benefits greatly the manufacturer in offering the right colors to the right customer group of its new product in order to achieve customer satisfaction. Correlation Coefficient can be found. According to that, we can identify the customer's behaviour.

In terms of limitations, this system takes time to train, particularly if there are a lot of different colors (data). This is because we design the ANN by using a ''supervised learning method'' that allows a more accurate and appropriate result with a large scale of data. If the amount of data is small, we may design the ANN by using an ''unsupervised learning method'' which uses a different set of equations. This paper initially aims to assess customer satisfaction on colors, and design therefore, the model itself is feasible and applicable to some other type of business.

**Authors**

**P.Isakki alias Devi** receieved B.Sc and M.C.A Degree from Madurai Kamaraj University in 1997 and 2000. She has received M.Phil Degree from Bharadidasan University in 2008. She is working as a Assistant Professor in MCA Department of Guru Nanak College, India. She has 10 years of teaching experience. She is pursuing Ph.D in Vels University, India. Her Research area is Data Mining for Customer Relationship Management.

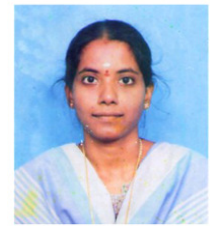

**Dr.S.P.Rajagopalan** received M.Sc from IIT Madrs, M.Phil and Ph.D from Madras University, Chennai, India. He is working as a Professor (Emeritus) in School of Computer Science & Engineering of M.G.R.University, Chennai. He has 40 years of teaching experience. He has published 4 Books. He has about 100 publications in International Journals and National Journals. His special fields of interest include Quantitative Techniques, Data Processing and Project Management, Management Information System, Programming Languages, Simulation, Text generation, Cryptography and Data Mining.

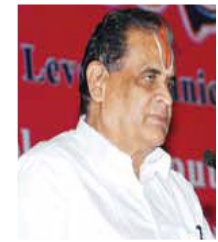